# An Approach to Model Interest for Planetary Rover through DSmT

Matteo Ceriotti,[1] Massimiliano Vasile[2]
*University of Glasgow, Glasgow, G12 8QQ, United Kingdom*

Giovanni Giardini,[3]
*Texas A&M University, College Station, Texas 77843-3141*

and
Mauro Massari[4]
*Politecnico di Milano, Milano, 20156, Italy*

**In this paper, we propose an approach for assigning an interest level to the goals of a planetary rover. Assigning an interest level to goals, allows the rover to autonomously transform and reallocate the goals. The interest level is defined by data-fusing payload and navigation information. The fusion yields an "interest map", that quantifies the level of interest of each area around the rover. In this way the planner can choose the most interesting scientific objectives to be analyzed, with limited human intervention, and reallocates its goals autonomously. The Dezert-Smarandache Theory of Plausible and Paradoxical Reasoning was used for information fusion: this theory allows dealing with vague and conflicting data. In particular, it allows us to directly model the behaviour of the scientists that have to evaluate the relevance of a particular set of goals. The paper shows an application of the proposed approach to the generation of a reliable interest map.**

---

[1] Ph.D. Candidate, Department of Aerospace Engineering, James Watt Building (South), Glasgow, United Kingdom, AIAA Member.
[2] Senior Lecturer, Department of Aerospace Engineering, James Watt Building (South), Glasgow, United Kingdom, AIAA Senior Member.
[3] TEES PostDoctoral Research Associate, 743D H.R. Bright Building, Ross Street - 3141 TAMU, College Station, Texas, USA.
[4] Assistant Professor, Department of Aerospace Engineering, Via La Masa 34, Milano, Italy.

## Nomenclature

| | | |
|---|---|---|
| $2^\Theta$ | = | power set of the frame $\Theta$ |
| $A, B, C$ | = | generic elements of the power set or hyper-power set |
| $Bel$ | = | belief function |
| $D^\Theta$ | = | hyper-power set of the frame $\Theta$ |
| $m$ | = | general basic belief number (gbba) or basic probability assignment (bpa) |
| $m_1, m_2$ | = | gbba or bpa assigned by each expert |
| $m_{12}$ | = | combined gbba or bpa |
| $m'$ | = | bpa on the refined frame |
| $I$ | = | interesting hypothesis |
| $NI$ | = | not-interesting hypothesis |
| $Pl$ | = | plausibility function |
| $u$ | = | uncertainty value |
| $X, Y, Z$ | = | generic elements of the refined power set |
| $\delta$ | = | discount due to uncertainty |
| $\theta$ | = | element (possible event) of a given frame of discernment |
| $\Theta$ | = | frame of discernment |
| $\Theta_{ref}$ | = | refined frame of discernment |

## I. Introduction

Based on the experience gathered with past Mars robotic missions, a number of future space missions envisage the use of robots for the exploration of distant planets [1]. All of them have strong scientific requirements but the poor knowledge of the environment where the robots will operate, makes the definition of specific goals dependent on contingent events and observations. If the allocation the goals is performed entirely on the ground, the robot will have to wait for new instructions every time a new, unforeseen event occurs or a new set of scientific data is available.

Therefore, it would be desirable to have an autonomous system able to make decisions not only on how to reach a given set of goals but also on which mission goals to select. Furthermore, the persistency of a mission goal may lead the system to repeatedly re-plan in order to meet the goal though the goal is unreachable or has lost its original importance. Goal transformation or goal reallocation is an important feature required in dynamic and rapidly changing environments but can become extremely important also in poorly known environments or when exploration and discovery are the main drivers of a mission [2]. For example, assume that, for a mission to Mars, a set of observations from space is used to define a set of goals for a planetary rover. During the mission, however, the rover may find that the goals are unreachable (e.g. if the goal was to collect a sample of a specific rock, the rock could be unreachable) or not interesting anymore (e.g. a different rock may display more interesting features). Then, the ground control team, together with the scientific community, would have to decide what to do. While the ground control team is devising a new plan and a new set of goals the rover would remain idle waiting for instructions. In order to avoid this waiting time, the idea is to adjust mission goals of the planner in addition to the adjustment of the plans themselves. Previous works on goal transformation addressed terrestrial or military applications [2, 3], and did not include the scientific data coming from the payload in the reallocation process.

In this work, we propose the autonomous generation or reallocation of given mission goals in order to maximise mission return. The aim is to have the most rewarding sequence of goals or the addition, deletion, modification of goals depending on contingent events or discoveries. Payload information is integrated in the planning process in order to make the rover mimicking the behaviour of scientists. Goals are generated, modified or reallocated in order to maximize the overall scientific return of the mission. A family of plans is then generated for each set of ordered goals and the most reliable feasible plan of the most interesting set of goals is executed. Reliability is taken into account, together with interest, in the process of choosing the plan to be executed [4]. The planner and the goal transformation algorithm are part of a multilayer autonomous system called Wisdom. The Wisdom system is a non-deterministic, deliberative-reactive system for rover autonomy in harsh, unknown environments. The system was developed and implemented on a six wheeled prototype rover (named Nausicaa), as part of a study for the development of advanced systems for space autonomy [5].

In this paper, we present specifically the approach used in Wisdom to generate an interest value through the data fusion of navigation and payload data for an autonomous planetary rover. The definition of an interest value avoids wild goal sequences for which only an empty set of actions is feasible (a plan with no steps), since only goals that are interesting for the mission can be generated or transformed. In Wisdom, goals are extracted from a pool of high level conceptual directives and are organized into a sequence by using the STRIPS paradigm for planning [6]. Briefly, goals are distributed in a logical sequence from an initial goal to a final one with preconditions and postconditions, but are not scheduled unless the time is explicitly part of a goal (e.g. reach a given location in a given time). The sequence can be adjusted during execution and is qualified according to the total level of interest of all the goals. The definition of a pool of high level directives limits the set of goals to those for which the autonomous system was designed but avoids the persistency of unreachable goals.

Previous attempts to model vague concepts such as interest or curiosity for autonomous agents can be found in the work of Schmidhuber [7], who proposed the use of a co-evolutionary algorithm to evolve curiosity in an artificial intelligence system. In this case, however, there is no specific use of instruments or any mission-specific measurements or data to support the decision-making process. Instead, in this paper, a full exploitation of scientific data is proposed in order to build an interest map of the surroundings. Pieces of scientific data from different sources are fused with navigation one to yield a single value for each point on the map. The map, then, evolves during the mission depending on the available observations.

In general terms, *data fusion* is the use of independent and/or redundant ancillary data from various sources to improve the data already available. Wald formally defined data fusion as: "A formal framework in which are expressed the means and tools for the alliance of data originating from different sources. It aims at obtaining information of greater quality" [8]. Here we understand data fusion as a way to combined information from different sources in order to obtain a single unambiguous value, useful to make decisions on the interest of a particular set of goals.

The combination of scientific and navigation data requires the fusion of pieces of information coming from physically different sensors. Each sensor measures a different parameter, has its own characteristics, reliability and uncertainty on measurements. Moreover, if each instrument is interpreted as a scientist expressing an opinion, we can associate to each data set an interest level with associated uncertainty. This would mimic the process performed on ground when a new set of scientific data is available. The data fusion process is then required to collect all the different pieces of information, with associated uncertainty, and combine them together [9].

In order to fuse data from the sensors and find the most interesting areas of the surrounding environment, the Dezert-Smarandache theory of plausible and paradoxical reasoning [10] was used. This theory has been successfully applied to many engineering problems, like the estimation of behaviour tendencies of a target [11], or the prediction of the land cover change [12]. In those works, it was proven that this modern theory overcomes the limitations of both fuzzy logic and evidence theory.

The main advantage of the paradoxical reasoning is that it allows dealing simultaneously with uncertain and paradoxical data from different, providing a solution even in the case of conflicting information. A conflict leads to a non-decidable situation that would put the rover into idle mode, waiting for instructions. The conflict could arise when different sources (different instruments) are assigning opposite interest values to the same area or when the navigation expert suggests avoiding an area that has a high level of interest. Conflicts on the ground would be resolved through a discussion among the scientists and the mission control team, leading to a new set of goals. An autonomous resolution of conflicts by the rover, would reduce the time spent to wait for instructions from the ground station.

In this paper, after a brief introduction to the theory of Plausible and Paradoxical Reasoning, the application to modelling interest for the Wisdom system is explained. The way of modelling interest fusing information from different sensors is described, and an application to a synthetic environment is shown. At the end, we will present a brief discussion about the possible use of Dempster-Shafer theory for the assignment of an interest. It should be noted that the key point of this work is not to propose a new theory of information fusion or to present the advantages of one theory over another. The key point is to propose an innovative way to assign a value of interest to mission goals for a planetary rover so that the goals can be autonomously adapted to contingent mission events.

## II. Plausible and Paradoxical Reasoning

The theory of plausible and paradoxical reasoning (or Dezert-Smarandache theory, DSmT [10]) is a generalisation of the Dempster-Shafer evidence theory [13], which is in turn a generalisation of the classical probability. The foundation of the DSmT is to abandon the rigid models of the previous theories, because for some fusion problems it is impossible to define or characterize the problem in terms of well-defined and precise and exclusive elements.

Given an experiment, the frame of discernment $\Theta = \{\theta_1, \theta_2, ..., \theta_n\}$ is the set of all possible events. The model on which the DSmT is based allows dealing with imprecise (or vague) notions and concepts between elements $\theta_i$ of the frame of discernment $\Theta$. The DSmT includes the possibility to deal with evidences arising from different sources of information which do not have access to absolute interpretation of the elements $\theta_i$ under consideration. This means that some events may also be overlapped and/or not well defined.

If $\Theta$ is the frame of discernment, we can define the space $D^\Theta$, called hyper-power set [14], as follows:

$$\varnothing, \theta_1, ..., \theta_n \in D^\Theta;$$
$$\forall A, B \in D^\Theta, (A \cup B) \in D^\Theta, (A \cap B) \in D^\Theta. \quad (1)$$

No other elements belong to $D^\Theta$, except those obtained by using rules in (1). Once $D^\Theta$ is defined, we can apply the map $m(\cdot): D^\Theta \to [0,1]$, called *general basic belief number*, or gbba [10], such that:

$$m(\varnothing) = 0,$$
$$\sum_{A \in D^\Theta} m(A) = 1.$$

A set of gbba, referred to the same frame of discernment $\Theta$, is called evidence.

This approach allows us to model any source that supports paradoxical (or intrinsically conflicting) information. The theory of Dezert-Smarandache defines a rule of combination for intrinsically conflicting and/or uncertain independent sources. If two experts give their opinions in terms of bodies of evidence $m_1$ and $m_2$, their combination is given by:

$$m_{12}(A) = \sum_{\substack{B,C \in D^\Theta \\ B \cap C = A}} m_1(B) m_2(C), \quad \forall A \in D^\Theta. \tag{2}$$

Note that this rule is commutative and associative and requires no normalization procedure. Moreover, it can manage the paradoxical information without any other assumption, thus overtaking some limitations of other probability theories – like the evidence theory – in which the frame of discernment shall be based on a set of exhaustive and exclusive elements.

All the pieces of evidence in Eq. (2) are then used to give two uncertainty values: the *belief* and the *plausibility*:

$$\begin{aligned} Bel(A) &= \sum_{B \in D^\Theta | B \subseteq A} m(B); \\ Pl(A) &= \sum_{B \in D^\Theta | B \cap A \neq \varnothing} m(B). \end{aligned} \tag{3}$$

The belief of an event $A$ is the sum of all the prepositions that totally agree with event $A$, while plausibility sums up all the prepositions that agree with $A$ totally or partially. An estimation through classical probability theory would fall in the interval defined by the values of belief and plausibility.

### III. Modelling Interest for a Planetary Rover

The high level of autonomy required to a planetary rover demands for the ability to choose the mission goals, without human intervention, once high level mission objectives are defined, in order to maximise the scientific return of the mission. These objectives, such as "look for water" or "look for traces of life", do not identify exactly where to go and which experiments to perform. The rover should be able to uniquely define what is interesting, by means of the information gathered during the mission, and make decisions without waiting for instructions from the ground station. The collected pieces of information can be incomplete and uncertain. In particular, the Wisdom system uses different sensors to obtain the pieces of evidence required to make a decision. Each instrument plays the role of a scientist or of a ground control specialist. DSmT is used to model the following situation: each scientist (or specialist) expresses an opinion on the interest of a given object or portion of the surrounding area; the scientist admits no uncertainty but the one that comes from the instruments. On the other hand, every scientist leaves some margin for discussion, accepting the existence of opposite opinions.

A. Modelling of sensor information

Nausicaa, the rover used to test the Wisdom system, is equipped with an infrared camera (the scientific payload) and two optical navigation cameras that give a stereographic view of the surrounding environment (the navigation module). The optical stereo images are used to generate an elevation map of the ground (called Digital Elevation Map or DEM). The DEM is a matrix containing the height of the corresponding point on the ground.

The DEM can be a partial reconstruction of the surroundings. Some parts of the terrain may not be in sight, because hidden by other parts (e.g. rocks or hills), and thus it is not possible to have any information about them. Furthermore, the algorithm can fail to determine the height of some points, especially if the image quality is poor. For these reasons, a second matrix is stored together with the DEM: it contains the uncertainty on the elevation of each point in the DEM. Values are between 0 and 1, where the former means total certainty on the elevation.

Besides giving information on the elevation of the ground, optical images provide information on the texture of objects and surfaces. A texture map is then created by associating to each point in view an integer value identifying a specific material. Since this information might not be accurate or the image could be poor, a map of uncertainty is associated to the texture map.

The payload mounted on Nausicaa generates a thermal map of the environment. This map is analogous to the DEM, but contains the temperature of each visible point. An uncertainty map is then associated to the thermal map, in order to take into account partial information due to occultation or the measurement noise of the infrared sensor. The final step consists of fusing the data of the three maps, to generate a single one: the interest map.

### B. Definition of the Interest Map

The interest map is a matrix in which each element represents the belief that a particular spot on the ground is interesting. A frame of discernment $\Theta = \{I, NI\}$ was defined, where $I$ is the hypothesis *interesting* and *NI* is the hypothesis *not-interesting*. Interest is a vague concept and is subjective in nature. The associated hyper-power set is defined as $D^\Theta = \{\emptyset, I, NI, I \cup NI, I \cap NI\}$, and gbbas are assigned to the interesting and not interesting hypotheses, but also to:

- $I \cup NI$: uncertain hypothesis. Represents the amount of ignorance, or the lack of knowledge of the expert which is dealing with the gbba assignment. The expert assigns evidence to this hypothesis when the uncertainty on the data is high, due for example to distance, error on the sensor, or even lack of data.

- $I \cap NI$: paradoxical hypothesis. This is the case in which two distinct scientists disagree on the interest level of a particular area. One of the scientists, according to the readings of his instruments, assigns a very high gbba to the interesting hypothesis while the other assigns a very high gbba to the not interesting hypothesis.

Note that in the classical probability theory, these two additional hypotheses do not exist. Furthermore, the difference between the uncertain and the paradoxical cases is that the former expresses uncertainty due to lack of knowledge or information, while the latter does not claim any ignorance, but the possibility that both hypotheses could be true at the same time.

As a consequence, the two associated hypotheses are vague, can be overlapped, and can not be considered as mutually exclusive. The various pieces of information can be conflicting and highly uncertain. These types of information can be effectively handled through DSmT since it can manage conflicts among various experts and provides a single rule of combination.

The interest map is created point by point, by fusing all the available pieces of information (or evidence that a point is interesting or not) about each one of the maps as summarized in Fig. 1. A set of independent experts (the instruments) creates the bodies of evidence that will be fused. For each point on the map the expert has to express an opinion on whether the point is interesting or not based on some evidence. The opinion is expressed by assigning gbba to each point on the map. The evidence comes form the readings of the navigation and scientific instruments. In particular, three experts were created, one for each map. The gbba that each expert assigns to a point on the map depends on the scientific objectives of the mission and on the available measurements. The measured values are compared against the values in a reference look-up table (the tables for the three experts can be found in Table 1 to Table 3). For example, in this work, we assume that the expert associated to the DEM is interested in sharp edges and in the lateral surface of the rocks since they are easily accessible. Thus, it assigns much gbba to the interesting case and little to the not interesting case, when the value of the gradient of the DEM is high, and vice-versa (Table 1). In addition, for some values of the gradient, gbba is also assigned to the paradoxical case. This is done not because of lack of knowledge of the roughness of the terrain (in which case, gbba is assigned to the uncertain hypothesis), but because the value of the gradient alone would not be sufficient to completely define whether an area is interesting or not. Assigning gbba to the paradoxical case allows for the integration of the opinions of other experts even if they are conflicting with the one of the DEM expert.

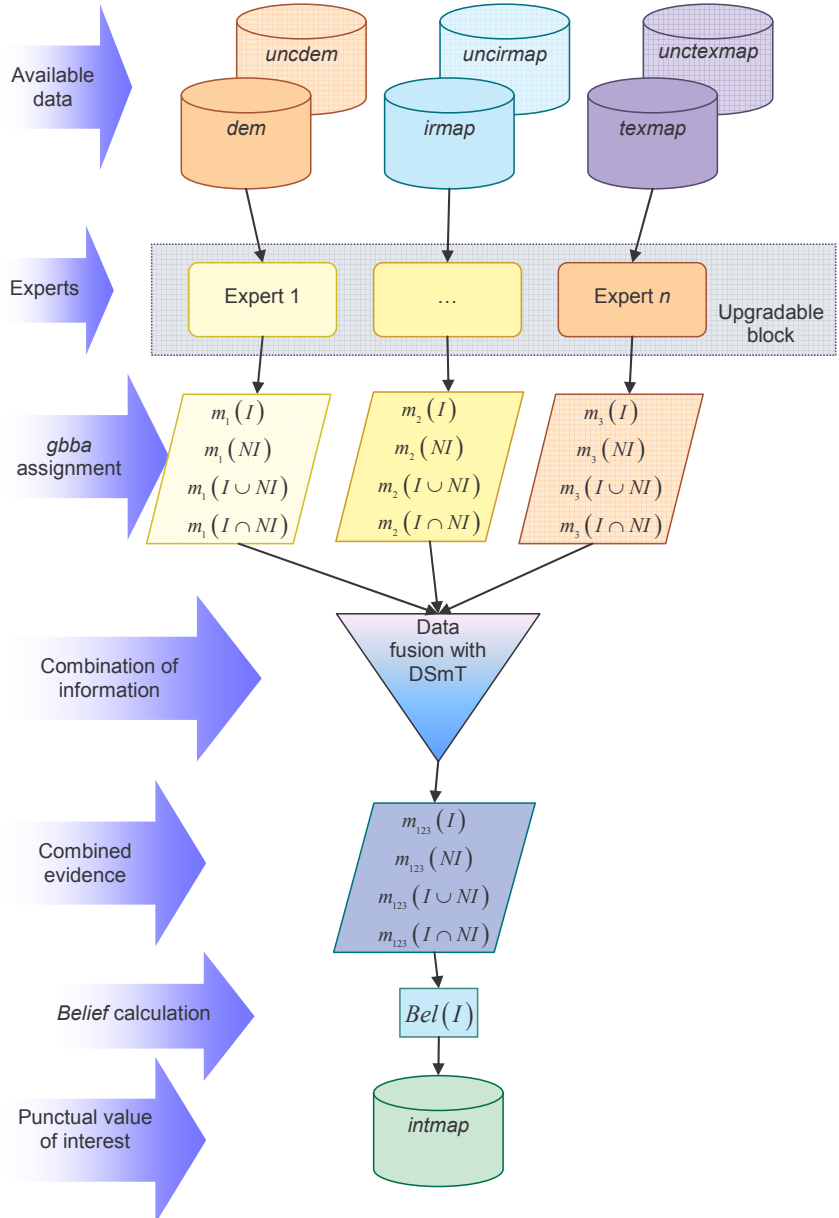

**Fig. 1. Diagram of the procedure to create the interest map.**

In the same way, the temperature expert assigns interest to some temperatures (Table 2), and the texture expert assigns interest to some specific textures (Table 3). As before, gbba is assigned to the paradoxical hypothesis when the values associated to temperature and texture cannot be used to completely establish whether the point is interesting or not.

**Table 1. Table for the DEM expert. Non-dimensional units have been used.**

| Modulus of the gradient of the DEM | $m(I \cap NI)$ | $m(NI)$ | $m(I)$ |
|---|---|---|---|
| $[0,1)$ | 0.20 | 0.80 | 0 |
| $[1,3)$ | 0.30 | 0.60 | 0.10 |
| $[3,5)$ | 0.10 | 0.10 | 0.80 |
| $[5,7)$ | 0.15 | 0.05 | 0.80 |
| $[7,9)$ | 0.05 | 0.05 | 0.90 |
| $[9,+\infty)$ | 0.05 | 0 | 0.95 |

**Table 2. Table for the temperature map expert. Non-dimensional units have been used.**

| Temperature | $m(I \cap NI)$ | $m(NI)$ | $m(I)$ |
|---|---|---|---|
| $[0,20)$ | 0.20 | 0.80 | 0 |
| $[20,40)$ | 0.40 | 0.50 | 0.10 |
| $[40,60)$ | 0.05 | 0 | 0.95 |
| $[60,80)$ | 0.15 | 0.05 | 0.80 |
| $[80,100)$ | 0.05 | 0.05 | 0.90 |

**Table 3. Table for the texture expert.**

| Texture | $m(I \cap NI)$ | $m(NI)$ | $m(I)$ |
|---|---|---|---|
| Texture not in database | 0.20 | 0.80 | 0 |
| 1 | 0.30 | 0.60 | 0.10 |
| 2 | 0.10 | 0.10 | 0.80 |
| 3 | 0.15 | 0.05 | 0.80 |
| 4 | 0.05 | 0 | 0.95 |

At first no gbba is assigned to the uncertain hypothesis $I \cup NI$; subsequently, each expert redistributes part of the basic probability associated to the hypothesis $I \cap NI$, $NI$, $I$ to the hypothesis $I \cup NI$. The gbba are redistributed proportionally to the value $u$ of the corresponding uncertainty map associated to each expert map, by using the following classical discounting procedure:

$$\left. \begin{array}{l} \delta(i) \leftarrow m(i) \cdot u \\ m(i) \leftarrow m(i) - \delta(i) \\ m(I \cup NI) \leftarrow m(I \cup NI) + \delta(i) \end{array} \right\} \quad i = I \cap NI,\ NI,\ I \quad (4)$$

The value of $u$ depends on the characteristics of the sensor (e.g. measurement errors) In this work, uncertainty maps will be simulated in order to provide a variety of test cases for the data fusion process. Therefore, the value of *u* will not be chosen to reproduce the actual measurements but just to test the proposed methodology. Note that, if the instruments are ideal and no uncertainty in their measurements is present, no mass is assigned to the hypothesis $I \cup NI$.

The assignment process presented in Eq. (4) is applied to each point on the DEM. Given the three sets of evidence by each expert, the general combination rule for paradoxical sources of DSmT is applied, and the combined evidence is computed. The following step is to compute the belief in the hypothesis *interesting*, $Bel(I)$. This value gives a pessimistic estimation (lower boundary) of the probability of that point to be actually interesting. Therefore, the interest map will contain, for each point on the DEM, the belief that point is interesting, according to the high level mission goals. The planner will then give more importance to those areas that are more likely to be interesting, and will reallocate the goals in order to maximise the cumulative value of interest with the highest reliability.

In the following section we will present how each maps are generated and how the belief is computed for a specific test case.

## IV. Simulation

The proposed approach was initially tested in a simulated environment. A synthetic landscape was generated inserting typical features like rocks with different textures and slopes with different gradients. The algorithm was run simulating the behaviour of the two navigation and the infrared cameras mounted on Nausicaa.

The aim of this sample test case was to generate an interest map that was consistent with the simulated features. The result was then used by the planner [5] to generate a set of mission goals in order to visit only the spots that are considered to be the most rewarding in terms of science.

The synthetic landscape, represented in Fig. 2, was converted into a DEM. The x-y plane in the figure represents an ideal horizontal plane, while z is the elevation of each point of the terrain with respect to this plane. Non-dimensional units for length and temperatures have been used. Assuming that the rover is in the centre of the map, and the height of the camera from the ground is 40 units, it has been possible to calculate whether each point of the map was in sight of the camera or not (Fig. 3).

As explained above, the module that generates the DEM provides also an uncertainty map based on visibility (partial information about the landscape) and on the intrinsic measurement errors of the digital cameras. The uncertainty map is initially created with values of zero (point in sight, no uncertainty on its elevation) or one (hidden point, no information about its elevation). Then, the uncertainty due to errors of recognition of the disparity maps are simulated by introducing a noise component, with a value in the interval $[0, 0.2]$. The resulting uncertainty map is represented in Fig. 4.

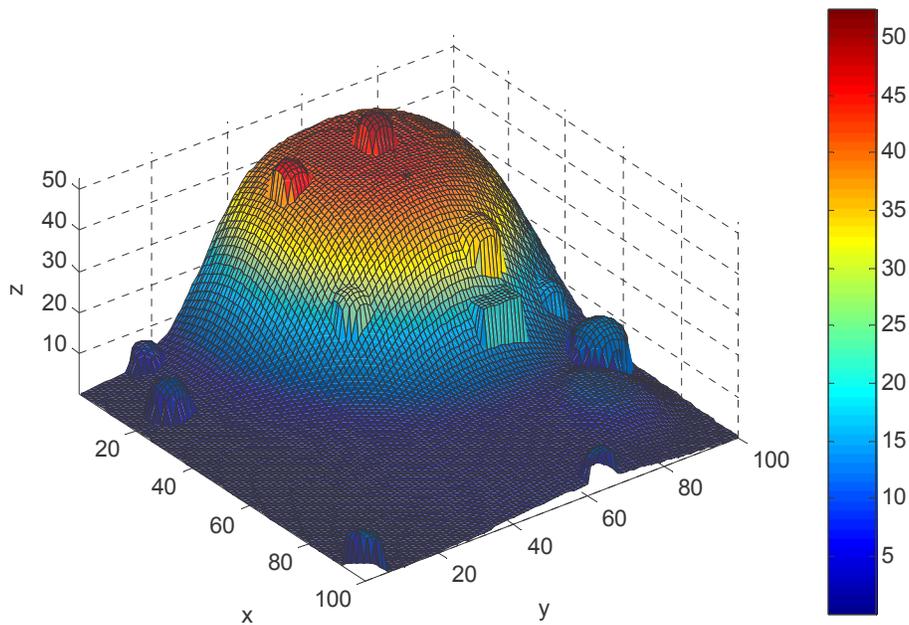

**Fig. 2. DEM of the synthetic landscape: bumped features represent rocks.**

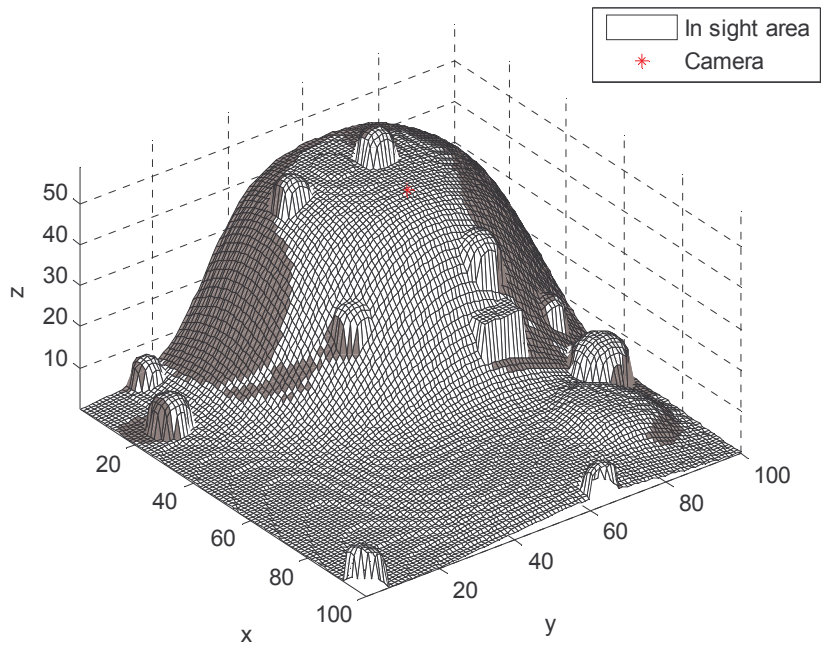

**Fig. 3. Visibility map superimposed on the DEM: in dark grey, surfaces that are not in sight. The camera is in the middle of the map, at a high of 40 units from the ground.**

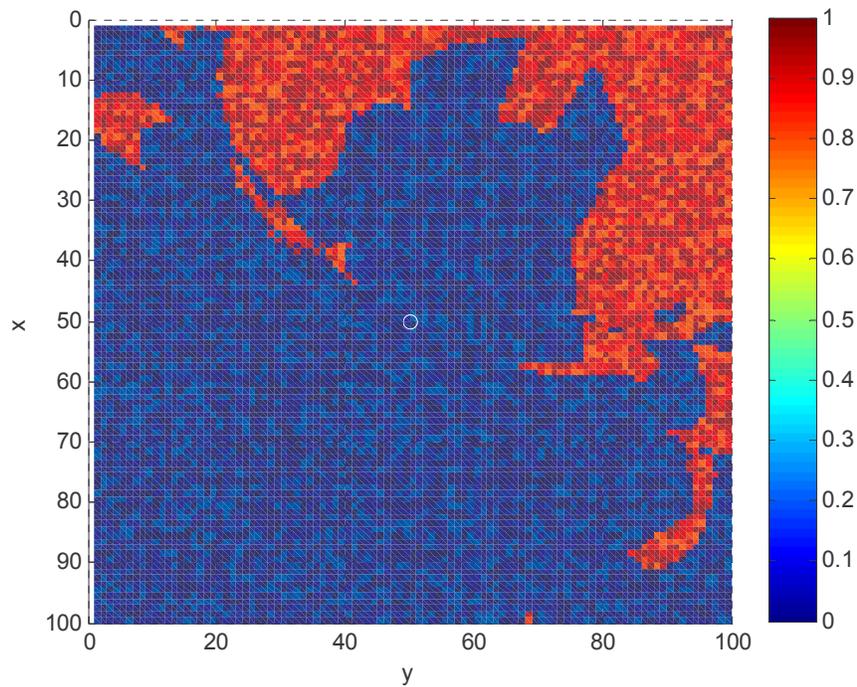

**Fig. 4. The uncertainty map associated to the DEM.**

The expert that creates the evidence from the DEM first computes the map of the gradient of the terrain, starting from its elevation; then, it assigns high interest to the points which have a high gradient, and low interest to other points (Table 1). In Fig. 5 there is a representation of the absolute value of the gradient of the DEM, as computed by the corresponding expert.

The virtual infrared map contains the temperature of the corresponding point on the DEM. The expert associated to the infrared camera assigns high levels of interest to hot areas. Fig. 6, shows the temperature distribution in the virtual environment: the whole terrain as an average temperature below 5 (in the non-dimensional units of temperature) which correspond to a cold terrain, apart from single circular hot area.

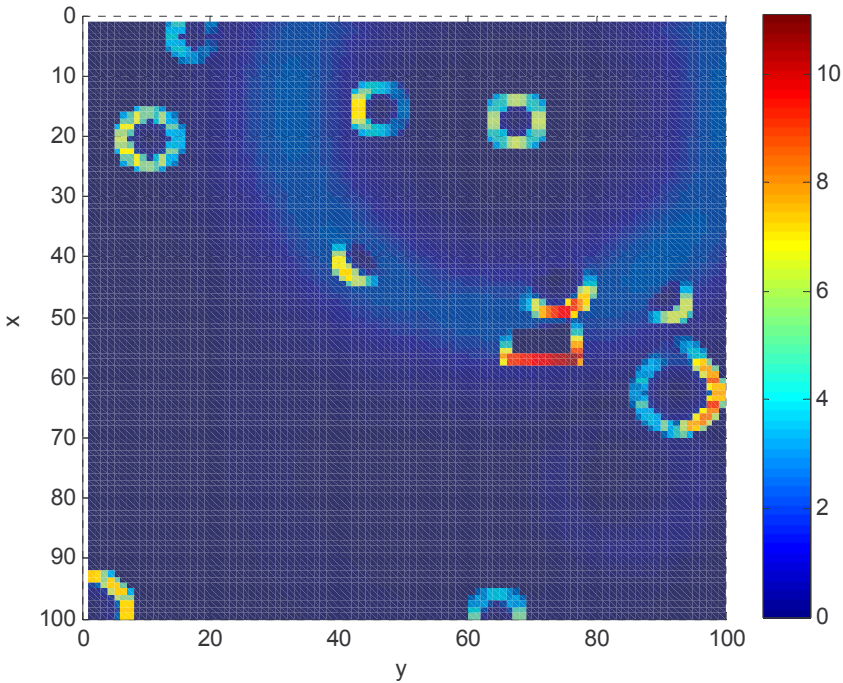

**Fig. 5. Representation of the absolute value of the gradient of the DEM.**

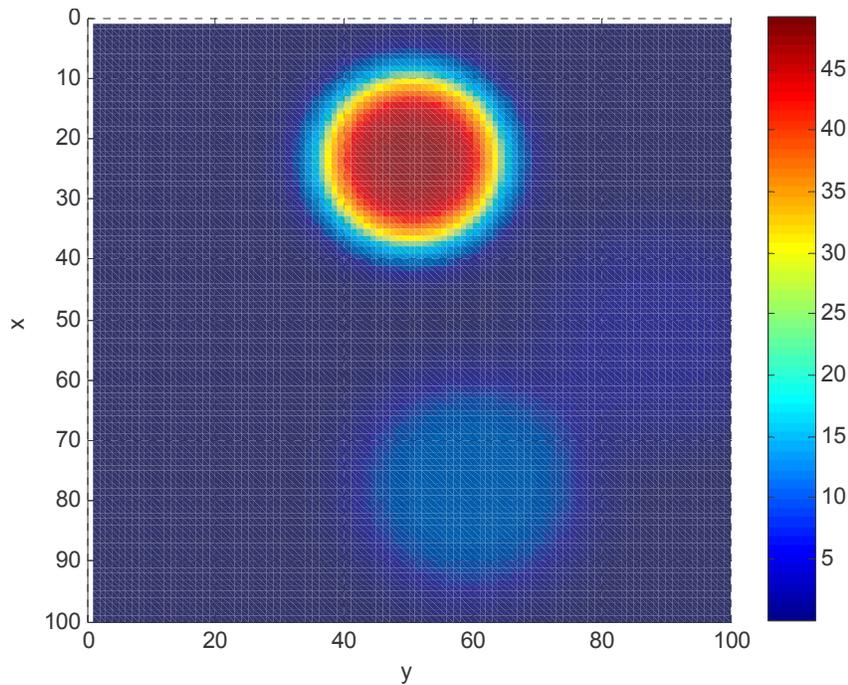

**Fig. 6. The infrared map.**

The texture distribution is represented in Fig. **7**: four different patterns have been considered, each one corresponding to one colour in the figure. The reference textures with their associated level of interest are stored in a database onboard. The expert of this map assigns the gbba according to the reference values in Table 3: it was assumed texture 4 (coloured in brown in Fig. 7) has the a greatest probability to be interesting for this particular mission.

The experts associated to the texture and infrared maps generate the corresponding uncertainty maps in a similar fashion as the expert of the DEM: they check for visibility of each point and surface in the map. In fact, if the infrared image and optical image are captured simultaneously, without moving the rover, the unknown areas must be the same. However, this yields the same level of uncertainty for the same points on all the three maps. Therefore it was assumed that the uncertainties for the infrared map grows linearly from the bottom end of the map to the upper end of the map, while the uncertainty on the texture grows linearly from the right end to the left end of the map, as shown in Fig. 8. Note that this assumption has no particular physical meaning, but it allows us to have areas with very different and mixed levels of uncertainties, thus testing properly the proposed data fusion framework. A different distribution of uncertainty, though producing different values, does not change the significance of the results presented in this paper. As stated above, in a real case, the uncertainty map would depend on the properties of the instruments and on the level of confidence of the scientists in their own judgement.

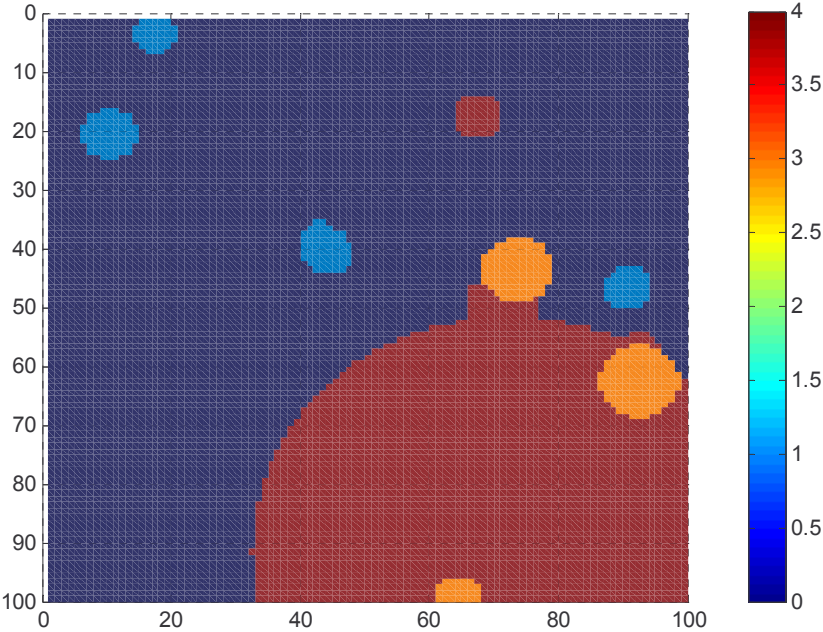

Fig. 7. The texture map: each colour is associated to a different texture in the database.

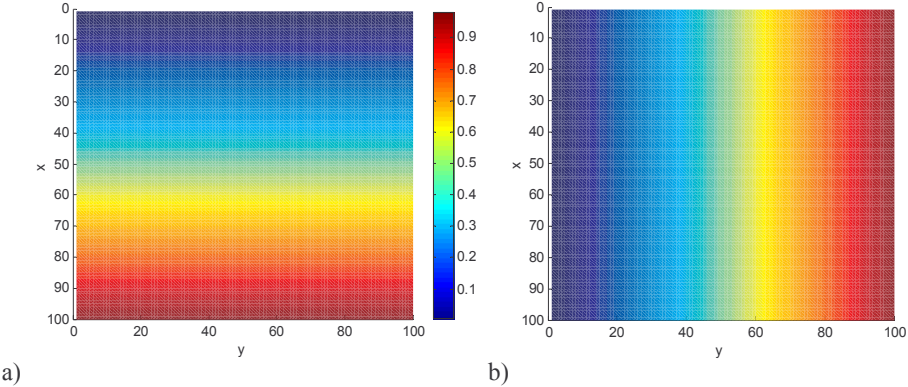

Fig. 8. The uncertainty associated to the infrared map (a) and to the texture map (b).

Given the maps and the experts, the result of the fusion process, as explained in par. III.B, is the interest map shown in Fig. 9. The value associated to each point in the map represents the belief that the point is interesting.

The areas identified by the letters A, B, C, D, E, F, G, H and I in Fig. 9, corresponding to rock borders, are marked as very interesting because of the high gradient value. It shall be noted that only the parts in sight of the cameras are interesting (this is particularly noticeable in the case of spots B, C, D and G). Where the rock is hidden, the gradient is high, but its unreliability is high, as well; thus, the assignment from the expert is uncertain and the associated belief is low.

The circular area identified with letter L is considered interesting mainly by the expert of the infrared map but its visibility is high as well as its reliability. In fact, Fig. 6 shows that the temperature is high in that area and Fig. 8a shows that for that area, the infrared map has a low uncertainty value; thus the information it gives is considered to be very reliable.

The small area with letter M is the most interesting of the whole map, with a value close to one. This is due to the synergy between the DEM and the infrared experts: both have certain information, and the gradient and the temperature are very high.

The sudden change in the level of interest on area N is a consequence of the discontinuity of the soil texture, as can be seen in Fig. 7. Looking at the map, starting from the area N, and moving right, the degree of interest gradually decreases: this is because the texture information is gradually less reliable on the right part of the map, as can be seen in Fig. 8b. Notice how both the infrared and the DEM expert regarded this area as not interesting but both the DEM and the texture experts stated that the reliability of what observed was good while the infrared stated the opposite. Nonetheless, the fused reliability of the texture and of the DEM maps supports the hypothesis that this area is worth a visit and is safe enough; as a consequence the associated belief is moderately high. Finally a three dimensional representation of the interest map superimposed onto the DEM can be seen in Fig. 10.

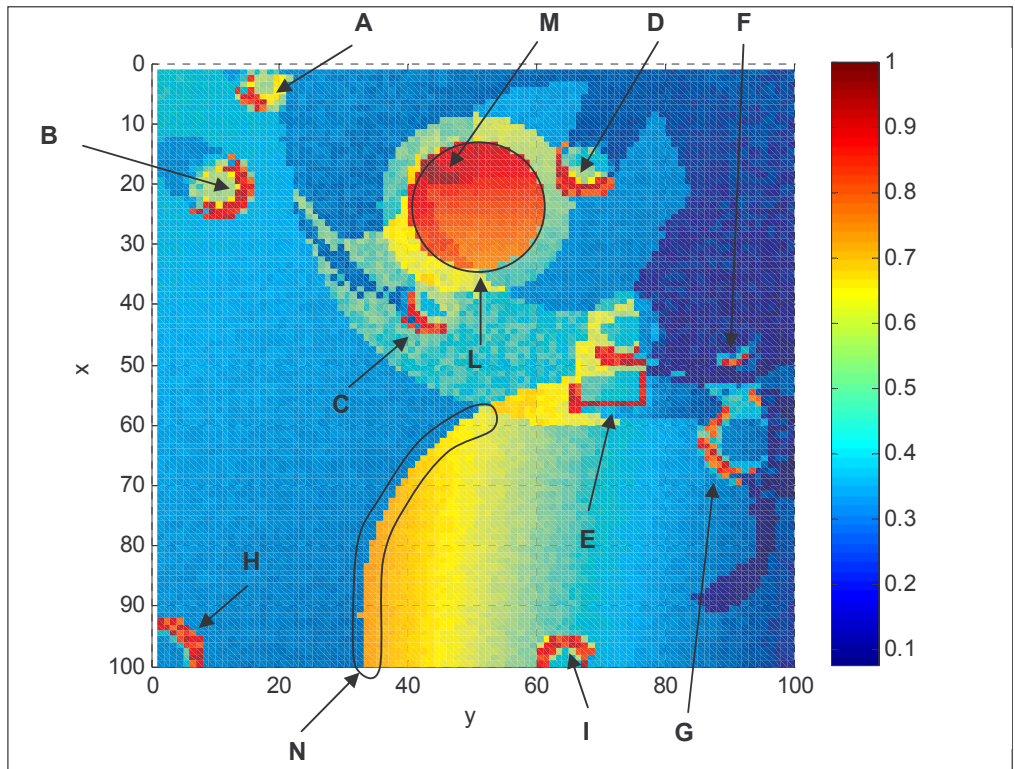

**Fig. 9.** Interest map: different colours represent different values of *Bel(I)*.

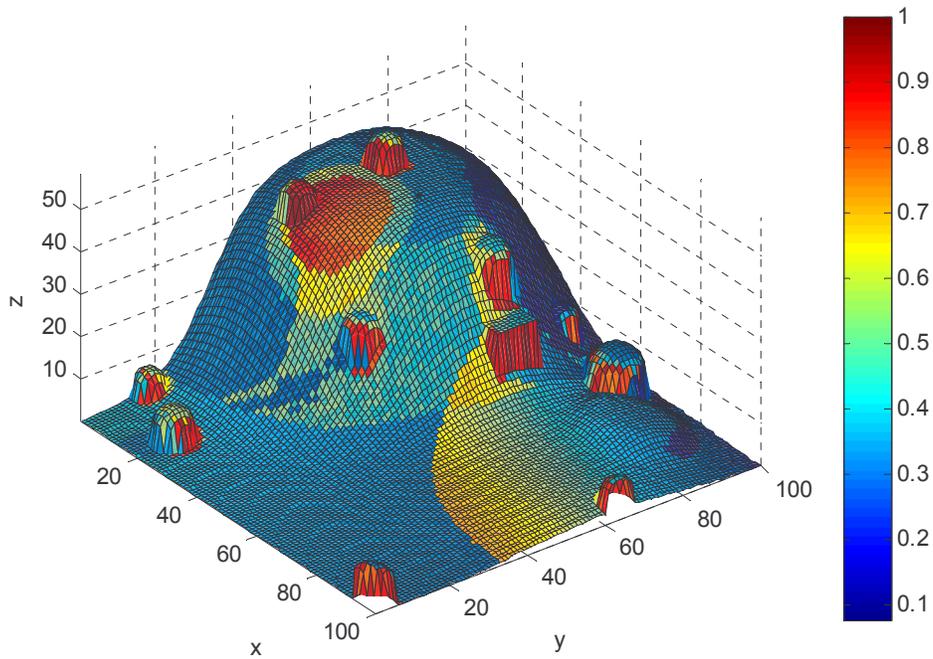

**Fig. 10.** Interest map superimposed on the DEM.

## A. On the use of DST for the generation of the Interest Map

The DSmT can be considered as an extension of the Dempster-Shafer Theory of Evidence (DST), from which it was derived. In fact, the DST is a particular case of the DSmT, in which all the sets of a given frame of discernment are disjoint (i.e., $\forall A, B \in \Theta, A \neq B \rightarrow A \cap B = \emptyset$). As a consequence, the set of possible hypotheses for a frame of discernment $\Theta = \{\theta_1, \theta_2\}$ is its power set $2^\Theta = \{\emptyset, A, B, A \cup B\}$. As for the DSmT, we have $m(\emptyset) = 0$ and $\sum_{A \in 2^\Theta} m(A) = 1$. In literature, the function $m(\cdot)$ is generally called *basic probability assignment* (bpa), when referred to the DST framework.

There are several different rules for combining bodies of evidence from different experts under this framework. The classical Dempster rule, which is associative and commutative, fuses the bpa $m_1$ and $m_2$ of two experts referred to the same frame of discernment in the following way:

$$m_{12}(A) = \frac{\sum_{\substack{B,C \in 2^\Theta \\ B \cap C = A}} m_1(B) m_2(C)}{1 - \sum_{\substack{B,C \in 2^\Theta \\ B \cap C = \emptyset}} m_1(B) m_2(C)}, \quad \forall A \in 2^\Theta \tag{5}$$

The Belief and Plausibility functions are computed in the same way as in the DSmT, that is using (3), given that the power set $2^\Theta$ shall be considered.

The different behaviour of the two theories is evident when conflicting bpas are given by the experts. In particular, the famous Zadeh's example [10] highlights the counter-intuitive results which the DST can lead to, while the DSmT is able to solve the contradiction in the sources of information quite easily, thank to the presence of the paradoxical hypothesis.

A simple case that brings to quite different results is when the assignments of two different sources are given, as in Table 4. In this case, the evidence of the two experts is almost totally conflicting, with a small uncertainty: this situation can happen, for example, when the terrain is flat (then not interesting for the DEM expert) but the texture is very interesting

The fusion through the DST, according to Eq. (5), leads to the combined bpa shown in the first column of Table 5. The DST combination rule assigns the same amount of evidence to both the hypotheses *I* and *NI*. In this framework, the value of $Bel(I)$ is the same as $m(I)$. In essence, the DST states that the point has the same probability of being interesting or not interesting, which does not allows the rover to take a decision on whether to investigate that point or not. On the other hand, the DSmT assigns most of the evidence to the paradoxical hypothesis $I \cap NI$, which is contributing in the value of $Bel(I)$.

**Table 4. Example of conflicting bodies of evidence for two different experts.**

|  | Expert 1 | Expert 2 |
|---|---|---|
| $m_i(I)$ | 0.99 | 0 |
| $m_i(NI)$ | 0 | 0.99 |
| $m_i(I \cup NI)$ | 0.01 | 0.01 |

**Table 5. Combined evidence and Belief according to DST and DSmT, for evidence provided by the two experts in Table 4.**

|  | DST | DSmT |
|---|---|---|
| $m_{12}(I)$ | 0.4975 | 0.0099 |
| $m_{12}(NI)$ | 0.4975 | 0.0099 |
| $m_{12}(I \cup NI)$ | 0.005 | 0.0001 |
| $m_{12}(I \cap NI)$ | - | 0.9801 |
| $Bel(I)$ | 0.4975 | 0.99 |

To show the different results in fusing the data using either the DST or the DSmT, let us consider the border of the rock D. As an example, we take the point (67, 20): for this point, we have the values for the gradient of DEM, texture and temperature listed in Table 6, with corresponding uncertainties. According to these values, the consequent bpas (or gbba) are also shown in the same table. The result of the combination through the DSmT is shown in Table 7. In conclusion, according to the DSmT, the point should be highly interesting, as the belief of the *I* hypothesis is close to one.

The use of the DST, instead, leads to a different result. The DST associative rule can be applied to the same point, but considering that in the DST framework, all the sets are disjoint, so $I \cap NI = \varnothing$, it would make no sense to assign bpa to this case. We decided here to reassign the gbba of the hypothesis $I \cap NI$ to the hypothesis $I \cup NI$, as in

Table 8, since a conflict of opinions would lead to a stall in the decision making process, analogous to a lack of information. Note that, for this case, a different choice of the bpa re-assignment would not change substantially the result obtained with the DST.

Applying the DST combination rule, we obtain the evidence in Table 9. Then we can state that, using the DST, the belief in the interesting hypothesis is significantly lower than for the DSmT. The border of the rock will not be a primary objective to analyse for the rover in this case.

**Table 6. Values of the three maps at point (67, 20), uncertainties, and corresponding assignments made by the experts.**

|  | Value | Uncertainty | $m(I \cap NI)$ | $m(NI)$ | $m(I)$ | $m(I \cup NI)$ |
|---|---|---|---|---|---|---|
| Gradient of the DEM | 6.088 | 0 | 0.15 | 0.05 | 0.8 | 0 |
| Texture | 4 | 0.66 | 0.017 | 0 | 0.323 | 0.66 |
| Temperature | 10.23 | 0.19 | 0.162 | 0.648 | 0 | 0.19 |

**Table 7. Combined evidence and Belief using the DSmT combination rule, for bodies of evidence given in Table 6.**

|  | Combined evidence |
|---|---|
| $m(I \cap NI)$ | 0.82293004 |
| $m(NI)$ | 0.02765400 |
| $m(I)$ | 0.14941600 |
| $m(I \cup NI)$ | 0 |
| $Bel(I)$ | 0.97234607 |

**Table 8. Re-assignment of the gbba of the paradoxical hypothesis to the uncertain hypothesis.**

|  | $m(NI)$ | $m(I)$ | $m(I \cup NI)$ |
|---|---|---|---|
| Gradient of DEM | 0.05 | 0.8 | 0 + 0.15 |
| Texture | 0 | 0.323 | 0.66 + 0.017 |
| Temperature | 0.648 | 0 | 0.19 + 0.162 |

**Table 9. Combined evidence and Belief using the Dempster combination rule, for bodies of evidence given in Table 8**

|  | Combined evidence |
|---|---|
| $m(NI)$ | 0.2296 |
| $m(I)$ | 0.6881 |
| $m(I \cup NI)$ | 0.0824 |
| $Bel(I)$ | 0.6881 |

1. Refined frame of discernment

If the frame of discernment is refined in the following way $\Theta_{ref} = \{I \cap NI, I/(I \cap NI), NI/(I \cap NI)\}$ then we can apply DST and obtain a result equivalent to the one computed using DSmT. Given the new refined frame of discernment, the power set is:

$$2^{\Theta_{ref}} = \{\emptyset, X, Y, Z, X \cup Y, X \cup Z, Y \cup Z, X \cup Y \cup Z\}$$

where:

$$X = I \cap NI$$
$$Y = I \cup (I \cap NI)$$
$$Z = NI \cup (I \cap NI)$$

Let us denote with prime the bpas referred to the refined frame of discernment. If we assign the bpas for each generic expert $i$ in the following way:

$$m'_i(X) = m_i(I \cap NI)$$
$$m'_i(X \cup Y) = m_i(I)$$
$$m'_i(X \cup Z) = m_i(NI) \quad (6)$$
$$m'_i(X \cup Y \cup Z) = m_i(I \cup NI)$$
$$m'_i\left(A \in 2^{\Theta_{ref}}, A \neq X, X \cup Y, X \cup Z, X \cup Y \cup Z\right) = 0$$

and the DST combination rule (5) is applied, we have that:

$$1 - \sum_{\substack{B,C \in 2^{\Theta_{ref}} \\ B \cap C = \emptyset}} m'_1(B) m'_2(C) = 1 \quad (7)$$

Computing the bpa, for example for $m'_{12}(X)$, we obtain:

$$\begin{aligned} m'_{12}(X) = &\ m'_1(X)m'_2(X) + m'_1(X \cup Y)m'_2(X \cup Z) + m'_1(X \cup Z)m'_2(X \cup Y) + \\ &\ m'_1(X)m'_2(X \cup Y) + m'_1(X)m'_2(X \cup Z) + m'_1(X)m'_2(X \cup Y \cup Z) + \\ &\ m'_1(X \cup Y)m'_2(X) + m'_1(X \cup Z)m'_2(X) + m'_1(X \cup Y \cup Z)m'_2(X) \end{aligned} \quad (8)$$

On the other hand, applying the DSmT combination rule (2) to the standard frame $\Theta = \{I, NI\}$, we obtain for $m_{12}(I \cap NI)$:

$$\begin{aligned}
m_{12}(I \cap NI) = &\, m_1(I \cap NI)m_2(I \cap NI) + m_1(I)m_2(NI) + m_1(NI)m_2(I) + \\
&\, m_1(I \cap NI)m_2(I) + m_1(I \cap NI)m_2(NI) + m_1(I \cap NI)m_2(I \cup NI) + \\
&\, m_1(I)m_2(I \cap NI) + m_1(NI)m_2(I \cap NI) + m_1(I \cup NI)m_2(I \cap NI)
\end{aligned} \quad (9)$$

Eq. (8) and Eq. (9) are equivalent and return the same value. Therefore, the fusion obtained using the DST with the refined frame of discernment, and the one obtained with the original model and DSmT are identical. The same happens for $m'_{12}(X \cup Y \cup Z)$, $m'_{12}(X \cup Z)$, $m'_{12}(X)$.

Note that, the refinement of the frame of discernment would require a probability assignment to the hypotheses $I/(I \cap NI)$ and $NI/(I \cap NI)$ that have little physical meaning and are not intuitive. Therefore, though DST can be used to define the interest map, DSmT offers a more direct definition and treatment of the two hypotheses $I$ and $NI$ without the need for an artificial redefinition of the frame of discernment. Furthermore, it should be noted that DSmT allows the direct treatment of a case in which a source is totally sure about its assignment and therefore cannot assign any probability to the hypothesis $I \cup NI$. In this case assigning a probability to the hypothesis $I \cap NI$ would correspond to allowing some room for discussion and opposite opinions as mentioned above.

## V. Conclusion

In this paper, an algorithm for the definition of the level of interest of mission goals for a planetary rover was presented. By fusing navigation data and payload data (infrared camera in this specific case), the rover was endowed with the capability to autonomously assign a level of interest to mission goals. Te interest level allows the rover to prioritise, reallocate and choose the most appropriate set of goals depending on contingent situations. The modern theory of Plausible and Paradoxical Reasoning was used to generate an interest map by which the rover can reallocate its goals autonomously in order to maximize the scientific return of the mission. The theory gives the possibility of dealing with vague quantities, like the degree of interest of an object. In particular, the advantage of DSmT is the possibility to directly assign a level of interest to hypothesis $I$ and $NI$ for each point of the DEM, leaving room for potential disagreements among the scientists or between the scientists and the ground control team.

The results showed that the proposed approach is suitable to uniquely identify the interesting zones, given the high level scientific goals of the mission. The goals can be easily modified or tuned, by changing the experts used into the data fusion process.

## Acknowledgments

This work is a part of the Wisdom project, for the development of a system for rover autonomy. Wisdom was developed under the ESA ITI contract 18693/04/NL/MV at Politecnico di Milano, when all the authors were working or studying there. The authors would like to thank Luc Joudrier of the robotic division of ESA/ESTEC. A special thank also to Flavio Fusco, Beatrice Midollini and Fabrizio Cappelli of Galileo Avionica, Lester Waugh and Ronan Wall of Astrium UK, for the industrial support and for the precious advices.

The authors would also like to thank Dr. Jean Dezert and Dr. Florentin Smarandache for their precious suggestions and ideas on how to apply DSmT and DST to our case.

## References

[1]  Baglioni, P., Fisackerly, R., Gardini, B., Gianfiglio, G., Pradier, A. L., et al., "The Mars Exploration Plans of Esa: The Exomars Mission and the Preparatory Activities for an International Mars Sample Return Mission," *IEEE Robotics and Automation Magazine*, Vol. 13, No. 2, 2006, pp. 83-89.

[2]  Stentz, A. and Hebert, M., "A Complete Navigation System for Goal Acquisition in Unknown Environments," *Autonomous Robots*, Vol. 2, No. 2, 1995, pp. 127-145.

[3]  Cox, M. T. and Veloso, M. M., "Goal Transformations in Continuous Planning," *Proceedings of the 1998 AAAI Fall Symposium on Distributed Continual Planning*, edited by M.d. Jardins, AAAI Press / The MIT Press, Menlo Park, CA, 1998.

[4]  Williamson, M. and Hanks, S., "Optimal Planning with a Goal-Directed Utility Model," *Proceedings of the Second International Conference on AI Planning Systems*, edited by K.J. Hammond, American Association for Artificial Intelligence, 1994, pp. 176-181.

[5]   Vasile, M., Massari, M. and Giardini, G., "Wisdom - an Advanced Intelligent, Fault-Tolerant System for Autonomy in Risky Environments, Final Report," ESA ITI Contract 18693/04/NL/MV, 2005.

[6]   Veloso, M., Carbonell, J., Perez, A., Borrajo, D., Fink, E., et al., "Integrating Planning and Learning: The Prodigy Architecture," *Journal of Experimental & Theoretical Artificial Intelligence*, Vol. 7, No. 1, 1995, p. 81.

[7]   Schmidhuber, J., "Artificial Curiosity Based on Discovering Novel Algorithmic Predictability through Coevolution," *Proceedings of the Congress on Evolutionary Computation*, edited by P.J.A.a.Z.M.a.M.S.a.X.Y.a.A. Zalzala, Vol. 3, IEEE Press, Lugano, Switzerland, 1999, pp. 1612-1618.

[8]   Wald, L., "Some Terms of Reference in Data Fusion," *IEEE Transactions on Geoscience and Remote Sensing*, Vol. 37, No. 3 pt 1, 1999, pp. 1190-1193.

[9]   Ceriotti, M., Vasile, M., Giardini, G. and Massari, M., "A Non-Deterministic Planner for Planetary Rover Autonomy," *Proceedings of the AIAA/AAS Astrodynamics Specialist Conference and Exhibit*, Keystone, CO, USA, 2006.
doi: AIAA 2006-6679

[10]  Dezert, J., "Foundations for a New Theory of Plausible and Paradoxical Reasoning," *Information & Security, an International Journal, edited by Prof. Tzv. Semerdjiev, CLPP, Bulgarian Acad. of Sci.*, Vol. 9, 2002.

[11]  Tchamova, A., Semerdjiev, T. and Dezert, J., "Estimation of Target Behavior Tendencies Using Dezert-Smarandache Theory," *Proceedings of the Sixth International Conference of Information Fusion*, Vol. 2, 2003, pp. 1349-1356.

[12]  Corgne, S., Hubert-Moy, L., Dezert, J. and Mercier, G., "Land Cover Change Prediction with a New Theory of Plausible and Paradoxical Reasoning," *Proceedings of the Sixth International Conference of Information Fusion*, Vol. 2, 2003, pp. 1141-1148.

[13]  Shafer, G., *A Mathematical Theory of Evidence*, Princeton University Press, Princeton, 1976.

[14]    Dezert, J. and Smarandache, F., "On the Generation of Hyper-Power Sets for the Dsmt," *Proceedings of the Sixth International Conference of Information Fusion*, Vol. 2, 2003, pp. 1118-1125.